\pdfoutput=1
\documentclass[11pt]{article}

\usepackage{ACL2023}

\usepackage{times}
\usepackage{latexsym}

\usepackage[T1]{fontenc}
\usepackage[utf8]{inputenc}

\usepackage{microtype}

\usepackage{inconsolata}

\title{User Study for Improving Tools for Bible Translation}

\author{Joel Mathew, Ulf Hermjakob \\
        Information Sciences Institute, University of Southern California \\
        \texttt{\{joel, ulf\}@isi.edu}}
\begin{document}
\maketitle
\begin{abstract}
Technology has increasingly become an integral part of the Bible translation process. Over time, both the translation process and relevant technology have evolved greatly. More recently, the field of Natural Language Processing (NLP) has made great progress in solving some problems previously thought impenetrable. Through this study we endeavor to better understand and communicate about a segment of the current landscape of the Bible translation process as it relates to technology and identify pertinent issues. We conduct several interviews with individuals working in different levels of the Bible translation process from multiple organizations to identify gaps and bottlenecks where technology (including recent advances in AI) could potentially play a pivotal role in reducing translation time and improving overall quality.
\end{abstract}

\section{Introduction}
Technology has played a key role in accelerating the Bible translation process by automating and assisting with time-consuming tasks. Today, multiple software tools have been developed that cater to users from varied backgrounds and following different translation processes. It is assumed that Bible translation being done today will rely on some technology to support their work. Routine tasks like editing and search are well studied and implemented in various tools available to translators.

The field of Natural Language Processing (NLP) is a field within the study of Artificial Intelligence (AI) that focuses on solving problem that relate to languages and text. The past decade has witnessed multiple points of inflection that dramatically improved the capability of NLP systems. We also observe that popular software tools used on the field for Bible translation do not take advantage of newer technology. Though this could be attributed to the lack of expertise in the development teams, we think there is still a bigger problem of clearly identifying use-cases and recommending changes in the translation process to effectively employ newer technology.

Arguably, technical teams that have developed and released applications on the field would have completed some form of user study to identify pain-points and areas where such a software tool could help. The conclusions of such studies, however, maybe colored by the technology stack and translation process or organization familiar to these teams. Moreover, the authors are unaware of any concerted efforts to share and collaborate on such learnings.

We attempt to tackle these issues through this study and communicate our findings. We consider our work as an important first step towards better understanding and communicating about the problem space and thereby encourage further participation from the Bible translation community at large. Our contributions in this work are:
\begin{enumerate}
\item We conduct interviews with multiple individuals involved with Bible translation in different language projects.
\item We report our preliminary findings and make recommendations for further investigation.
\end{enumerate}

In the following sections we describe our interviewing process, discuss our observations and finally share our conclusions.

\section{Interviews}
Taking interviews is a quick and effective means to gather a breadth of information and identify important areas for further investigation. It maybe difficult to make the case that a study is thoroughly systematic if it only relies on interviews. Instead of making such a claim, we consider this study an important first step in the right direction.

Overall, we conduct 7 hour-long interviews involving 16 individuals over video-conferencing. We record\footnote{Only for internal circulation; cannot be shared.} the interviews and keep detailed notes of our interactions. Most of our interviewees work on Bible translation projects in India (Central, East and North) though a few work on projects in Nepal and East Asia\footnote{Place names have been obscured due to the sensitivity of the work.}. One consultant currently works on developing resources for translators in English for subsequent translation.

\begin{table}
\centering
\begin{tabular}{lc}
\hline
\textbf{Interviewee Role} & \textbf{Frequency}\\
\hline
Mother Tongue Translator (MTT) & 3\\
Quality Checker (QC) & 4\\
Consultant-in-Training (CIT) & 4\\
Consultant & 3\\
Project Manager/Coordinator (PM) & 2\\
\hline
\end{tabular}
\caption{Self-reported roles of interviewees surveyed for this study.}
\label{tab:accents}
\end{table}

Before the start of each interview, we re-shared our intent (as described in this document) and ensured anonimity of the interviewees. Moreover, we clarified our position as interested third-parties and that any candid feedback about their organizations or tools would be kept private (especially the meeting recordings). Though we kept the tone of the interviews casual so as to elicit genuine and open responses, we were largely guided by a Question Bank \ref{sec:questionbank} developed to keep us grounded to the topic at hand which we share as an appendix of this report.

\section{Observations and Discussion}
\subsection{Technology Friendliness}
It was observed that there were translation team members who were introduced to a computer (laptop) for the first time just three months prior to project start along with individuals who considered themselves experienced and technology savvy. There were also individuals educated until high-school working along with Ph.Ds, having decades of experience and expertise on the field.
Backgrounds of the translation team members can vary considerably. Any tool or technology that is designed to interact with translation teams would benefit from designing applications that are targeted and marketed for a subset of the users with specific backgrounds. One-size may not fit anyone well. Moreover, assuming most of the remaining languages yet to be translated are spoken by minorities (when compared to LWCs\footnote{Languages of wider communication}), there is a bigger demand for tools with simpler user interfaces (UIs) that do not assume significant background knowledge from the user.

\subsection{Improved Resources}
A frequent request was for better resources in related languages other than LWCs to quickly identify context and produce accurate translations. Consultant notes were reported to be helpful though sometimes were only available in English which limited their accessibility. Moreover, it was not unusual for the teams to look up relevant resources available in multiple other languages (based on their ability to read) to make translation decisions for the more difficult passages.

Finding accurate meaning and context for words or phrases (including figures-of-speech) unfamiliar to the translators was reported as difficult and as a time-sink. Dictionaries were helpful though seeing relevant pictures would be better (e.g. for the word ephod). There are, however, legitimate instances where this process could take time. Especially, if there did not exist an equivalent term in the target language. e.g. 'yeast' was unknown to one of the target language communities.

There is a need for a resource that lists cultural references in the Bible with their location and significance. e.g. Exchanging slippers to make a covenant in the book of Ruth. It was noted that having access to good resources empower the lingual church to take ownership and more responsibility of the translation process.

In terms of integration with tools, offline access was reported as important. Tools that unobtrusively show only the parts of the reference resources that are relevant to the verse/word/phrase at-hand without requiring the user to switch screens were considered more helpful. Being able to quickly see all the translations of an exact word or phrase in related languages (those known to the translator) would greatly speed-up the drafting process while improving quality. Automated script conversion could be used to make more resources accessible for the translator.

\subsection{Translation Consistency}
There did not seem to exist a good strategy around maintaining consistency across Biblical books. An effort is made by limiting the number of people working on a book at a time, reading through drafts repeatedly and using the same list of key Biblical terms (limited scope) across books. Even when checking based on the key Biblical terms list, the current tools did not automatically account for inflections and required the users to manually list all possible inflections (e.g. affixes) as separate entries which makes maintaining the list laborious and unwieldy.

We recommend using available NLP techniques to automatically flag inconsistent usage of terms based on context and inflections, just in time. Such a feature could also tackle another class of difficult-to-spot issues: inconsistent usage of tense and singular/plural form between the source and target text. Moreover, suggesting semantically similar words (e.g. God and Lord) during drafting was a stated need. Consultants seem to rely on their experience and knowledge to spot inconsistencies in the text and would appreciate a tool to aid them.

\subsection{Spell Checking}
Spell checking is done by manually reading over all words used in the drafted text and fixing incorrect spellings case-by-case. General guidance for translators is to develop the dictionary in the target language before starting the translation process though most teams do not seem to follow this advice.

Though the list of words are automatically generated in the tool, we see this as an area that can be improved using modern techniques: it should be possible to flag suspicious spellings during drafting without requiring users to do this post hoc (e.g. after drafting a verse). Moreover, current tools show spelling suggestions based on a manually curated list of accepted words (dictionary). This can be augmented using statistical techniques to incrementally improve the quality of the suggestions based on the availability of the drafted corpus.

\subsection{Bottleneck Stages}
Multiple translators reported that the initial 'Drafting' (including 'Peer Checking') and 'Exegesis' stages are the most time consuming since they involve deep thinking to keep the text faithful to the source. The time taken in subsequent stages were reported to be dependent on the quality of the output of these initial stages. A consultant noted that achieving clarity for the New Testament and both clarity and accuracy for the Old Testament take time.

We recommend incorporating machine translation suggestions during the drafting stage, especially at the word/phrase level. Though starting with a machine translated first draft could be useful, it was reported that these are counterproductive and slow the work down when the machine generated output is of poorer quality. Moreover, highly specific (at the word/phrase level) just-in-time reference resources for translators would improve the both speed and quality of the initial draft.

\subsection{Backtranslation}
As a translation project progresses the backtranslations become more important. This is especially true for consultants who usually rely on them to catch issues and assess accuracy. Having more than one version of the backtranslation was reported as being helpful to better grasp the text. The advances in Neural Machine Translation makes it possible today to bootstrap reasonable quality of backtranslations using limited training data. Integrating such capabilities into tools would speed up the review process.

\subsection{Fonts and Inputs}
Typing the text in the target language has been noted as a time-consuming task (e.g. selecting diacritics from a list, etc.), despite using software tools like Google Input Tools\footnote{https://www.google.com/inputtools} and Phonetic. Most of the translators did not learn to type in their scripts directly citing a sharp learning curve. Some tools did not handle complex fonts correctly (e.g. for search and replace). Also, it was noted that there is a tendency to use hard-copies of drafts for checking and revising even though software tools facilitate this directly on the computer. This points to the importance of features that generate quick readable/printable formats from the drafts.

\subsection{Tool Complexity}
Complex tools (in terms of user interface and features) discourage users from trying newer or more advanced features due to fear of losing or messing up the data. It was also noted that strict tool policies (e.g. parallel passages) can sometimes confuse users and cause them to make changes in the text just to make the tool happy rather than reasoning through the text. This may also point to the need of having just-in-time documentation and guides (demonstrations) for tools.

\subsection{Community Checking}
It was noted that there did not seem to be much structure for the 'Community Checking' stage. This could be improved by stating concrete requirements for the number and types of participants along with simple methodologies to elicit meaningful feedback from the community (listeners).

\subsection{Translation Progression}
It was observed that the general high-level progression for Bible translation is from New to Old Testament. Though there maybe exceptions, this information is useful in designing tools that leverage incremental learning. e.g. Automated machine translation suggestions that are intentionally biased to learn contextually similar words found both in the New and Old Testaments.

\subsection{Quality Checking: Mental Modal}
There was an interesting perspective shared by an experienced consultant which we believe is helpful in thinking about the role of automated techniques for quality checking Bible translations: it is possible to think of quality checking being comprised of objective and subjective parts. The former is based on fixed patterns that a machine could learn and where technology could provide automated tooling for. Whereas it should provide support for the latter which has to do with the specificities of the target culture and language. Moreover, it maybe argued that a significant portion of the quality checking process is objective since the original languages (Hebrew, Greek) remain constant.

\section{Conclusion}
We survey multiple individuals through interviews who are working in different roles in the Bible translation process from various parts of the world and share our observations and make relevant recommendations for tool developers. Overall, we find multiple areas where the current state of technology could significantly improve the speed and quality of the Bible translation process.

\appendix
\section{Appendix}
\subsection{Interview Question Bank}
\label{sec:questionbank}
\subsubsection{Personal}
\begin{itemize}
\item What is your name and where are you based?
\item Which project(s) do you work in?
\item What is your educational background?
\item What languages do you speak?
\item How did you get involved with Bible translation and for how long?
\end{itemize}

\subsubsection{Technical}
\begin{itemize}
\item Do you own a smartphone or use a computer outside of work? If so, for how long?
\item Which software applications do you use for work? On what device?
\item What are three things you like most about the applications you use?
\item What are the three biggest issues you faced in the applications?
\item What feature/app do you think would really help you speed up your work?
\item Do you have a wish list for the tools you use?
\item Have you used AutographaMT?
\begin{itemize}
\item If so, which language did you use it for? What was your experience like?
\item What could be done better?
\end{itemize}
\item Have you used Paratext? If so, did you need training for it- how long?
\begin{itemize}
\item How easy/difficult do you find it?
\item What 3 features do you like the most?
\item What are 3 things do you like the least?
\end{itemize}
\end{itemize}

\subsubsection{Translation Related}
\begin{itemize}
\item Explain in detail all the steps involved in your work.
\begin{itemize}
\item Include applications/tools you use and how you interact with them.
\item How much time does each step take? How can each step be done faster?
\end{itemize}
\item Explain the progression of the work and what the process is after the current step.
\item What are you most concerned about when working on each step?
\item What tools/improvements do you wish for in the current work you are doing?
\item Would you be willing to share your screen and show us an example of your work?
\item What do you enjoy most about this work? What do you dislike?
\end{itemize}

\subsubsection{Feedback}
\begin{itemize}
\item What were your thoughts when you were asked about this interview?
\item How could we have helped you be better prepared for this interview?
\item Would you be willing to do another interview, if required?
\item What did you like and what could have been done better about this interview?
\item Are there others you think who could also provide relevant information? Who and why?
\end{itemize}
\end{document}